\documentclass{article}

\usepackage{PRIMEarxiv}

\usepackage[utf8]{inputenc} 
\usepackage[T1]{fontenc}    
\usepackage{hyperref}       
\usepackage{url}            
\usepackage{booktabs}       
\usepackage{amsfonts}       
\usepackage{nicefrac}       
\usepackage{microtype}      
\usepackage{lipsum}
\usepackage{fancyhdr}       
\usepackage{graphicx}       
\graphicspath{{media/}}     

\usepackage{algorithm}
\usepackage{algorithmic}

\usepackage{xcolor}

\usepackage{multirow} 
\usepackage{graphicx}
\usepackage{rotating}
\usepackage{xcolor}
\usepackage{algorithm}
\usepackage{algorithmic}
\usepackage{amsmath}
\usepackage{amssymb}
\usepackage{booktabs}
\usepackage{adjustbox}
\usepackage{graphicx}
\usepackage{verbatim}
\usepackage{newfloat}
\usepackage{listings}

\title{TimeSieve: Extracting Temporal Dynamics via Information Bottleneck}

\author{
  Ninghui Feng${}^*$ \\
  HKUST(GZ) \\
  \texttt{ninghuifeng@hkust-gz.edu.cn} \\
   \And
 Songning Lai\thanks{The first two authors contributed equally to this work.} \\
  HKUST(GZ) \\
  Deep Interdisciplinary Intelligence Lab ($DI^2 Lab$)\\
  \texttt{songninglai@hkust-gz.edu.cn}\\
  \And
  Jiayu Yang\\
HKUST(GZ)\\
Deep Interdisciplinary Intelligence Lab ($DI^2 Lab$)\\
\texttt{jyang729@connect.hkust-gz.edu.cn} \\
\And
Fobao Zhou\\
HKUST(GZ)\\
\texttt{fobaozhou@hkust-gz.edu.cn} \\
\And
Zhenxiao Yin\\
HKUST(GZ)\\
\texttt{zyin368@connect.hkust-gz.edu.cn}\\
\And
  Hang Zhao\thanks{Correspondence to Hangzhao \{hangzhao@hkust-gz.edu.cn\}}\\
HKUST(GZ)\\
\texttt{hangzhao@hkust-gz.edu.cn} \\
\And
}

\begin{document}
\maketitle

\begin{abstract}
Time series forecasting has become an increasingly popular research area due to its critical applications in various real-world domains such as traffic management, weather prediction, and financial analysis.
Despite significant advancements, existing models face notable challenges, including the necessity of manual hyperparameter tuning for different datasets, and difficulty in effectively distinguishing signal from redundant features in data characterized by strong seasonality. 
These issues hinder the generalization and practical application of time series forecasting models.
To solve this issues, this paper propose an innovative time series forecasting model - \textbf{TimeSieve} designed to address these challenges. 
Our approach employs wavelet transforms to preprocess time series data, effectively capturing multi-scale features without the need for additional parameters or manual hyperparameter tuning. 
Additionally, this paper introduce the information bottleneck theory that filters out redundant features from both detail and approximation coefficients, retaining only the most predictive information. This combination reduces significantly improves the model's accuracy.
Extensive experiments demonstrate that our model outperforms existing state-of-the-art methods in most cases, achieving higher predictive accuracy and better generalization. Our results validate the effectiveness of our approach in addressing the key challenges in time series forecasting, paving the way for more reliable and efficient predictive models in practical applications.

\end{abstract}

\maketitle

\section{Introduction}
\label{intro}

The field of time series forecasting has experienced significant advancements and found widespread applications in various real-world engineering tasks, such as traffic management \cite{8344781}, pandemic spread \cite{kumar2020covid}, and financial analysis \cite{sezer2020financial}. From early models like GRU \cite{8053243} and LSTM \cite{greff2016lstm} to the more recent Transformer-based architectures \cite{liu2023nonstationary,wu2021autoformer}, a plethora of sophisticated models have emerged. For instance, TimesNet \cite{wu2023timesnet} and FreTS \cite{yi2023frequencydomain} approach data processing from a frequency domain perspective, while multi-scale models like Pyraformer \cite{liu2022pyraformer} and Scaleformer \cite{shabani2023scaleformer} leverage intricate multi-scale and cross-dimensional dependency structures to significantly enhance performance on complex forecasting tasks.

Howerver, existing models often require the introduction of additional parameters to capture multi-scale features, which increases computational complexity. Moreover, these models necessitate manual tuning of hyperparameters for different datasets, further complicating their application and limiting their generalization across diverse scenarios. More importantly, many models, including Autoformer \cite{wu2021autoformer} , struggle to effectively distinguish between signal and redundant features when dealing with data characterized by strong seasonality. Redundant features not only interferes with the learning process, leading to increased errors, but also obscures the true patterns within the data, making the predictions unreliable. This issue is particularly pronounced when traditional autocorrelation mechanisms break down due to redundant features interference, significantly diminishing the practical value of these models.

To address these challenges, we proposes an innovative time series forecasting model designed to significantly improve predictive accuracy and broaden applicability.
First, we employ wavelet transforms to preprocess the time series data using the Wavelet Decomposition Block (WDB) and the Wavelet Reconstruction Block (WRB). Wavelet transforms efficiently capture information from multiple scales, but this process may also introduce redundant features. To address this, we integrate the Information Filtering and Compression Block (IFCB), an information bottleneck module that compresses and extracts the most relevant features by filtering out redundant features from both detail and approximation coefficients. These components work together, significantly enhancing the model's performance.

Our main contributions can be summarized as follows:
  
\begin{itemize}
    \item This paper introduce WDB and WRB that efficiently capture multi-scale features in time series data without requiring manual hyperparameter tuning or the introduction of additional parameters. 
      
    \item This paper utilize an information-theoretic approach by incorporating an IFCB to filter out redundant features from the data. This ensures that only the most predictive features are retained, enhancing the model's accuracy.
      
    \item By integrating wavelet transforms for comprehensive feature extraction and the information bottleneck block for redundant features, our model achieves state-of-the-art performance, demonstrating superior predictive accuracy and generalization across diverse scenarios.
      
\end{itemize}


\section{Related Work}
\label{sec:related_work}

\subsection{Time Series Forecasting}
In recent years, the field of time series forecasting has advanced rapidly, giving rise to many efficient models. These models predominantly rely on multilayer perceptron (MLP) approaches like MSD-Mixer \cite{zhong2024multiscale}, TSMixer \cite{chen2023tsmixer}, DLinear \cite{zeng2022transformers}and FreTs \cite{yi2023frequencydomain}, along with Transformer-based techniques such as PatchTST \cite{nie2023time}, Autoformer, and Informer \cite{zhou2021informer}, to perform predictions using sophisticated data transformation and learning mechanisms.

Time series forecasting can be defined as follows: 
Given a historical sequence of observations \(\ X=\{x_{1},x_{2},\ldots,x_{T}\}\), where $T$ is the lookback window length, the goal is to predict future values \(\ Y=\{y_{T+1},y_{T+2},\ldots,y_{T+H}\} = F(X)\), where $H$ is the forecast window length and $F$ is the predictive model mapping the input sequence $X$ to the output sequence $Y$.
However, time series data often exhibit complex, multi-scale characteristics. Conventional feature extraction methods, like standard pooling, require additional parameter adjustments or rely on manually designed pooling sizes, increasing model complexity and limiting flexibility. Wavelet transform, a classical signal processing technique \cite{farge1992wavelet}, excels in multi-scale analysis. It decomposes signals into different frequency components, aiding in the understanding and prediction of intricate patterns.

We propose a novel approach that combines wavelet transform with modern machine learning techniques for time series forecasting. The wavelet transform preprocesses the input data, decomposing it into detail coefficients and approximation coefficients at different scales. We believe this method effectively captures comprehensive features from the time series data.

\subsection{Information Bottleneck Theory}

In recent years, with the growing intersection of information theory and deep learning research, many researchers have begun analyzing and optimizing neural networks from the perspective of information theory. In this trend, information bottleneck (IB) theory has gained particular attention \cite{tishby2015deep,shwartz2017opening}, finding wide application in emerging fields such as graph neural networks \cite{dai2023unified}, multimodal learning \cite{mai2022multimodal}, and recently in time series imputation \cite{choi2024conditional}. IB theory helps improve model performance by optimizing information flow and processing \cite{tishby2015deep}.

We believe that wavelet transformation can effectively decompose time series data into components that contain valuable information. However, these components also include a significant amount of redundant information. If not properly addressed, this redundant information can interfere with the model's learning performance and predictive accuracy. 

From the perspective of information theory, IB theory provides a mechanism to effectively retain critical information related to target variables while filtering out irrelevant or redundant information in the model's intermediate representations. This approach enhances the model's ability to capture key features. Furthermore, it strikes a delicate balance between compressing input information and maintaining predictive accuracy, ensuring that the model's output remains precise and concise by:

\begin{equation}
\begin{aligned}
& \max I(y;z) \\
&  \min I(x;z)\\
\end{aligned}
\end{equation}

\noindent where $I(y;z)$ represents the mutual information between the output $y$ and the intermediate representation $z$, and $I(x;z)$ represents the mutual information between the input $x $and the intermediate representation $z$. Maximizing $I(y;z)$ ensures that the most relevant information for predicting the output is retained, while minimizing $I(x;z)$ helps to compress the input information, filtering out redundant features.

Therefore, this work seeks to apply information-theoretic methods to time series forecasting. Unlike prior applications of the IB theory in deep learning, we do not simply apply it directly. Instead, we integrate it into the overall framework to specifically handle the componets obtained from the wavelet transformation. In this way, IB theory can effectively filter out redundant features from these coefficients while retaining key information, ultimately enhancing the model's predictive performance.

  
\section{TimeSieve}
\label{sec:TimeSieve}
\begin{figure*}[htbp]
  \centering
      
  \includegraphics[width=\linewidth]{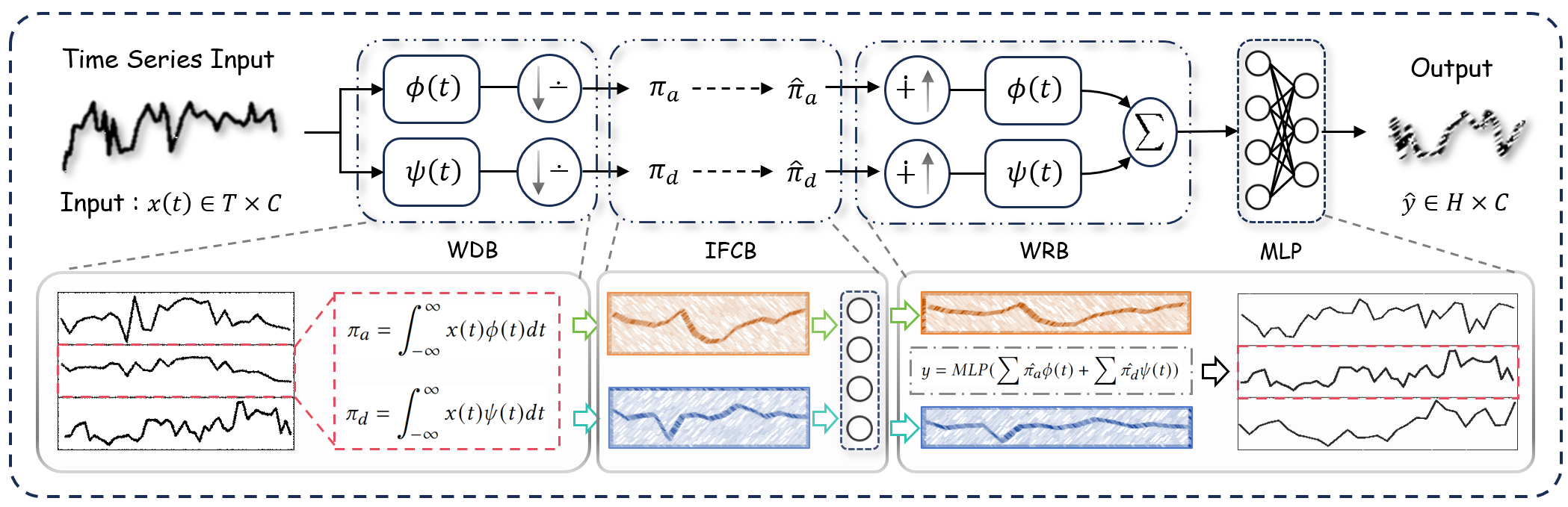}
  \caption{Overview of the TimeSieve Framework. The upper part illustrates the overall architecture of TimeSieve, which consists of WDB, WRB and IFCB. The input time series data, $T \times C$, is decomposed by WDB into coefficients $\pi_a$ and $\pi_d$, which are processed by IFCB. The WRB then reconstructs the data, followed by the predictor generating the corresponding forecast steps. The lower part provides a detailed visualization of the data flow within WDB and IFCB, with further visualization of IFCB available in Figure \ref{fig:IFCB}.}
  \label{fig:model}
      
\end{figure*}

\begin{figure}[tbp]
  \centering
  \includegraphics[width=0.4\linewidth]{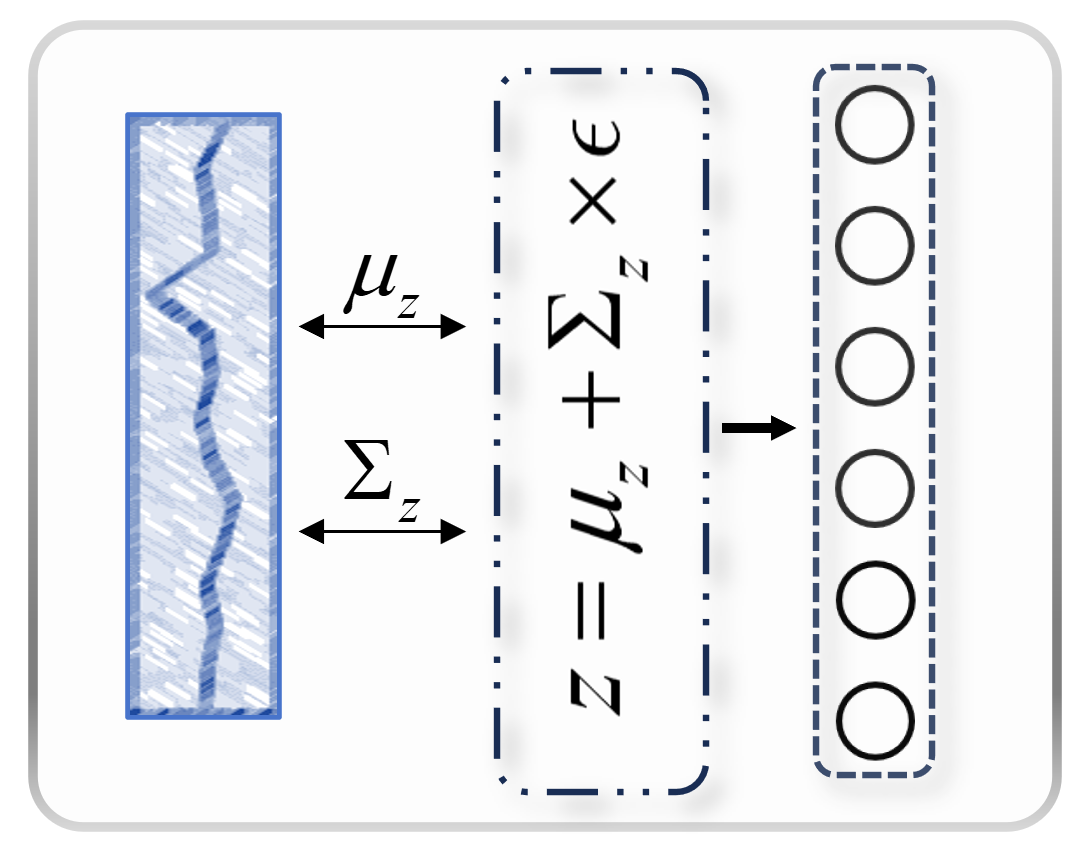}
  \caption{Overview of IFCB: This module consists of a reparameterization step and deep neural networks. Initially, the standard deviation and mean ($\mu_z, \Sigma_z$) of the data are computed. The reparameterization is then performed using the equation $z = \mu_z + \Sigma_z \times \epsilon$. The input coefficients are processed through the first neural network to obtain $\mu_z$ and $\Sigma_z$. The reparameterized $z$ is then fed into the second neural network to produce the reconstructed coefficients $\hat{\pi_i}$.}
      
  \label{fig:IFCB}
\end{figure}

In our study, we propose a novel time series forecasting model named \textbf{TimeSieve}. This model integrates the Information Filtering and Compression Block (IFCB) with wavelet transform method to enhance the accuracy of time series predictions. 

We explain the whole framework first, and we explain WDB, WRB and IFCB in more detail after that.

To effectively capture comprehensive feature information, we employ the Wavelet Decomposition Block (WDB). The WDB decomposes time series data into different frequency components, effectively extracting multi-scale information. Specifically, the wavelet transform can be represented by the following equations:

\begin{equation}
[\pi_a,\pi_d] = WDB(x(t))
\end{equation}

\noindent where $x(t)$ represents the input, the approximation coefficients $\pi_a$ represents the low-frequency trends, and the detail coefficients $\pi_d$ represents the high-frequency details, respectively.

However, the extracted multi-scale information may contain redundant features, which can adversely affect the model's learning performance and predictive accuracy. To address this issue, we introduce the IFCB to filter the information. Additionally, we employ residual connections to ensure completely keeping the valuable information during the filtering process. The IFCB optimizes the information flow, retaining critical information while filtering out irrelevant or redundant information. The IFCB could be shown as following equation:

\begin{equation}
\hat{\pi_a} = IFCB(\pi_a), \hat{\pi_d} = IFCB(\pi_d)
\end{equation}
    

\noindent where $\hat{\pi_a}$, $\hat{\pi_d}$ represents the filtered coefficients $\pi_a$, $\pi_d$.

After filtering with the IFCB, we apply the Wavelet Reconstruction Block (WRB) to reconstruct the processed data back into the time domain. This step ensures that the features at different scales are fully utilized. Finally, this paper use an MLP as the prediction layer to make the final forecast on the processed time series data, specifically adjusting the output to match the desired forecast length. The prediction step is formulated as follows:

\begin{equation}
Y = MLP(WRB(\hat{\pi_i})), i \in [a,d]
\end{equation}

The overall architecture of the TimeSieve model is illustrated in Figure \ref{fig:model}. The TimeSieve model leverages the combination of wavelet transform and IFCB to effectively handle the multi-scale characteristics present in time series data, thereby improving the model's predictive performance. This approach is motivated by the need to optimize information flow and enhance feature extraction, ensuring that our model can make accurate and reliable predictions across various applications.

The above is an overall description of TimeSieve. We will provide a detailed introduction to WDB, WRB, and IFCB in the following sections.

\noindent \textbf{WDB and WRB. }To effectively capture comprehensive feature information, we employ wavelet transformation to decompose time series data inspired by \cite{meyer2023time,nason2024leveraging}, allowing us to analyze the data across multiple scales. Wavelet transformation allows us to break down the data into various frequency components, enabling us to analyze and extract features across multiple temporal scales. This multi-scale analysis is crucial for identifying patterns and trends that occur at different time resolutions.

\noindent \textbf{WDB.} For the input time series $x(t) \in \mathbb{R}^{T\times C}$, where $T$ represents the length of the lookback window and $C$ denotes the number of variates, the wavelet transform decomposes the series into approximation coefficients $\pi_a$ and detail coefficients $\pi_d$. The approximation coefficients $\pi_a$ capture the general trend or smooth component of the series, representing low-frequency information. In contrast, the detail coefficients $\pi_d$ capture the fine-grained, high-frequency information, highlighting local fluctuations and details within the series. This decomposition can be expressed as:

\begin{equation}
\pi_i = \int x(t) \phi(t) dt, i \in [a,d]
\end{equation}
    

\noindent where $\phi(t)$ and $\psi(t)$ are the scaling function and wavelet function, respectively, defined as follows:

\begin{equation}
\phi(x) = \sum_{k=0}^{N-1} a_k \phi(2u - k), \psi(x) = \sum_{k=0}^{M-1} b_k \phi(2u - k)
\end{equation}
    

In the case of wavelet transform, $\phi(t)$ represents the low-pass filter, capturing the smooth, approximate part of the signal, and $\psi(t)$ represents the high-pass filter, capturing the detailed, high-frequency part of the signal. Here, $a_k$ and $b_k$ are the filter coefficients for the scaling and wavelet functions, respectively, $u$ is the variable over which the functions are defined, and $k$ is the index of the coefficients.

This decomposition allows the model to capture different characteristics of the data at various levels, thereby optimizing the understanding and prediction of time series dynamics, particularly in identifying local features within the series.

\noindent \textbf{IFCB. }Motivated by the need to effectively filter out redundant features while retaining critical information, we introduce IFCB to achieve this goal, inspired by the information bottleneck principle \cite{tishby2015deep,shwartz2017opening,liu2024timex}. In this section, we process the approximation coefficients $\pi_a$ and detail coefficients $\pi_d$ from the wavelet transform of the time series data, for convenience in subsequent explanations, we define $\pi_i $ and $i \in [a,d]$ to collectively represent $\pi_a$ and $\pi_d$. Additionally, We use $\hat{\pi_i}$ to represent the filtered form(s) of $\pi_i$. Initially, we define the input to the IFCB as $\pi_i \in \mathbb{R}^{T/2 \times C}$ and the output to the IFCB as $\hat{\pi_i} \in \mathbb{R}^{T/2 \times C}$.

In our model, both $\pi_i$ and $\hat{\pi_i}$ are treated as random variables, which given a joint distribution $p(i, \hat{i})$. The mutual information between these two random variables denoted as $I(\pi_i; \hat{\pi_i})$, quantifies the amount of information shared by $pi_i$ and $\hat{\pi_i}$ . Mutual information between joint distribution is defined as followed,

\begin{equation}
\begin{aligned}
I(\pi_i; \hat{\pi_i})&=D_{kL}[p(i, \hat i)\|p(i)p(\hat i)] = \sum_{i\in \mathcal{I},\hat{i}\in \hat{\mathcal{I}}}{p\left( i,\hat{i} \right) \log \left( \frac{p\left( i,\hat{i} \right)}{p\left( i \right) p\left( \hat{i} \right)} \right)}\\
&= \sum_{i\in \mathcal{I},\hat{i}\in \hat{\mathcal{I}}}{p\left( i,\hat{i} \right) \log \left( \frac{p\left( i|\hat{i} \right)}{p\left( i \right)} \right)} = H\left( \pi_i \right) -H\left( \pi_i | \hat{\pi_i} \right) 
\end{aligned}
\end{equation}

\noindent where $D_{kL}[p\|q]$ is the Kullback-Liebler divergence of the distributions $p$ and $q$, and $H(P)$ and $H(P|Q)$ are the entropy and conditional entropy of $P$ and $Q$, repectively.

We define the intermediate hidden layer random variable $Z$, which follows the Markov chain principles, where $\pi_i \rightarrow Z \rightarrow \hat{\pi_i}$, the mutual information satisfies $I(\pi_i; Z) \geq I(\pi_i; \hat{\pi_i})$. Although there is an inevitable information loss in the transmission to the middle layer $Z$, notice that $Z$ retains the most critical information for predicting $\hat{\pi_i}$ while filtering out irrelevant features.

This property makes the definition of objective of IFCB reasonable, which is to maximize the mutual information between $Z$ and $\hat{\pi_i}$ while minimizing the information between $Z$ and $\pi_i$:

\begin{equation}
\min (I(\pi_i; Z) - \beta I(Z; \hat{\pi_i}))
\end{equation}

\noindent where $\beta$ is a trade-off parameter, which determines the weight of the minimal information constraint in the optimization process.

The optimization is carried out within a deep neural network, which is designed to minimize the information between $Z$ and $\pi_i$ by adjusting the network weights and biases through backpropagation and suitable regularization techniques.

Then optimization is performed within a deep neural network. Assuming that $p(z|i)$ follows a Gaussian distribution, and the following derivation is established:

\begin{equation}
\begin{aligned}p(z|i) = \mathcal{N}(\mu(i;\theta_\mu), \Sigma(i;\theta_\Sigma)) = \mathcal{N}(\mu_z, \Sigma_z)\end{aligned}
\end{equation}

\noindent where $\theta_\mu$  and $\theta_\Sigma$  are the parameters of the networks that are optimized during training. 

The parameter updating process posing a new challenge due to the stochastic property of the gradient calculations. To address the challenge, the reparameterization trick is introduced. The reparameterization is formulaed by:

\begin{equation}
z=\mu_z+\Sigma_z\times\epsilon  
\end{equation}

\noindent where $\epsilon$ is sampled from a standard Gaussian distribution.

Given that the study focuses on a regression task, we define the conditional probability $q(\hat i|z)$:

\begin{equation}
q(\hat i|z)=e^{-||\hat i-D(z;\theta_c)||}+C
\end{equation}

\noindent where $D$ is a decoder function parameterized by $\theta_c$, and $C$ is a constant. This formulation measures the fit of the model's fitting by capturing the deviation of the predicted output $\hat i$ from the output of the decoder.

\noindent \textbf{WRB.} After the wavelet transformation, the detail and approximation coefficients are processed through IFCB to preserve the most predictive information. Subsequently, the inverse wavelet transform is applied to reconstruct the time series from the processed coefficients:

\begin{equation}
\hat{x}(t) = \sum \hat{\pi_a} \phi (t) + \sum \hat{\pi_d} \psi (t)
\end{equation}

\noindent where $\hat{\pi_a}$ and $\hat{\pi_d}$ are the filtered approximation and detail coefficients, respectively. Finally, the reconstructed time series is passed through a MLP for final prediction.

\begin{algorithm}
\vspace{-0pt}
\caption{Pseudo code of TimeSieve}
\begin{algorithmic}[t]
\REQUIRE Time series data $ x(t) = \{x_{1}, x_{2}, \ldots, x_{T}\} $, true labels $ Y=\{y_{T+1},y_{T+2},\ldots,y_{T+H}\} $
\ENSURE Predicted output $\hat{Y}$

\STATE \textbf{Initialization:} Initialize the model parameters, learning rate $\eta$, and total training epochs $E$

\STATE \textbf{Training:}
\FOR{$epoch \leftarrow 1$ \TO $E$}
        \STATE $\pi_a, \pi_d$ $\leftarrow$ WDB($X(t)$)
        
        \FOR{$i \in \{a, d\}$}
            \STATE $ \hat{\pi_i}, \mu_{z}, \Sigma_{z} \leftarrow \text{IFCB}(\pi_i) $
        \ENDFOR
        
        \STATE $ X_{\text{reconstructed}} \leftarrow $ WRB($\hat{\pi_a}, \hat{\pi_d}$)
        \STATE $ \hat{Y}_i \leftarrow \text{MLP}(X_{\text{reconstructed}}) $
        
        \STATE $\mathcal{L}_o \leftarrow \text{LossFunction}(\hat{Y}_i, Y_i)$
        
        \FOR{$i \in \{a, d\}$}
            \STATE $\mathcal{L}_{IB_I} \leftarrow D_{KL}[\mathcal{N}(\mu_{zI}, \Sigma_{zI}) \,||\, \mathcal{N}(0, I)] + D_{KL}[p(z_I) \,||\, p(z_I|i)]$
        \ENDFOR
        
        \STATE $\mathcal{L}_{IB} \leftarrow \mathcal{L}_{IB_A} + \mathcal{L}_{IB_D}$
        \STATE $\mathcal{L} \leftarrow \mathcal{L}_o + \mathcal{L}_{IB}$
        \STATE $\theta \leftarrow \theta - \eta \nabla_\theta \mathcal{L}$
\ENDFOR

\STATE \textbf{Inference:}
    \STATE $\pi_a, \pi_d$ $\leftarrow$ WDB($x(t)$)
    
    \FOR{$i \in \{a, d\}$}
        \STATE $ \hat{\pi_i}, \mu_{z}, \Sigma_{z} \leftarrow \text{IFCB}(\pi_i) $
    \ENDFOR
    
    \STATE $ X_{\text{reconstructed}} \leftarrow $ WRB($\hat{\pi_a}, \hat{\pi_d}$)
    \STATE $ \hat{Y}_i \leftarrow \text{MLP}(X_{\text{reconstructed}}) $

\RETURN final predictions $\hat{Y}$
\label{al:1}
\end{algorithmic}
\end{algorithm}
%


\noindent \textbf{Loss Funtion. }We now specify the model's loss function. This function is crucial for training the model to minimize prediction errors and optimize the distribution parameters effectively. The loss is composed of the original loss component and the IFCB loss, as shown below:

\begin{equation}
\begin{aligned}
\mathcal{L} = \mathcal{L}_o + \mathcal{L}_{IB} 
            = \mathcal{L}_o + D_{KL}[\mathcal{N}(\mu_z, \Sigma_z) \,||\, \mathcal{N}(0, I)] + D_{KL}[p(z) \,||\, p(z|i)]
\end{aligned}
\end{equation}
\vspace{00pt}

\noindent where $\mathcal{L}_o$ is the original prediction task loss, typically representing the error in regression predictions, and $\mathcal{L}_{IB}$ is the IFCB loss.

The overall architecture of IFCB is illustrated in Figure \ref{fig:IFCB} and pseudo code of TimeSieve is shown in Algorithm 1. 

\begin{table*}[htp]
    
  \centering
  \caption{Forecasting results with different forecast lengths H $\in$ $\{48, 96, 144, 192\}$ . We set the lookback length T = 2H. \textcolor[rgb]{ 1,  0,  0}{Red} indicates the best result, while \textcolor[rgb]{ 0,  .439,  .753}{Blue} refers to the second best.}
  \adjustbox{max size={0.85\textwidth}{\textheight}}{
    \begin{tabular}{c|ccccccccccccccc}
    \toprule
    \multicolumn{2}{c}{Models} & \multicolumn{2}{c}{TimeSieve} & \multicolumn{2}{c}{Koopa} & \multicolumn{2}{c}{PatchTST} & \multicolumn{2}{c}{DLinear} & \multicolumn{2}{c}{NSTformer} & \multicolumn{2}{c}{LingtTS} & \multicolumn{2}{c}{Autoformer} \\
    \midrule
    \multicolumn{2}{c}{Metric} & MAE   & MSE   & MAE   & MSE   & MAE   & MSE   & MAE   & MSE   & MAE   & MSE   & MAE   & MSE   & MAE   & MSE \\
    \midrule
    \multirow{4}[2]{*}{\begin{sideways}ETTh1\end{sideways}} & 48    & \textcolor[rgb]{ 1,  0,  0}{0.361 } & \textcolor[rgb]{ 1,  0,  0}{0.341 } & 0.385  & 0.364  & 0.375  & \textcolor[rgb]{ 0,  .439,  .753}{0.342 } & \textcolor[rgb]{ 0,  .439,  .753}{0.372 } & \textcolor[rgb]{ 0,  .439,  .753}{0.342 } & 0.465  & 0.614  & 0.406  & 0.404  & 0.432  & 0.678  \\
          & 96    & \textcolor[rgb]{ 1,  0,  0}{0.384 } & \textcolor[rgb]{ 1,  0,  0}{0.376 } & 0.411  & 0.406  & 0.395  & \textcolor[rgb]{ 0,  .439,  .753}{0.377 } & \textcolor[rgb]{ 0,  .439,  .753}{0.393 } & 0.380  & 0.498  & 0.653  & 0.431  & 0.435  & 0.496  & 0.578  \\
          & 144   & \textcolor[rgb]{ 1,  0,  0}{0.397 } & \textcolor[rgb]{ 1,  0,  0}{0.393 } & 0.426  & 0.424  & 0.412  & \textcolor[rgb]{ 0,  .439,  .753}{0.394 } & \textcolor[rgb]{ 0,  .439,  .753}{0.401 } & \textcolor[rgb]{ 0,  .439,  .753}{0.394 } & 0.536  & 0.602  & 0.442  & 0.453  & 0.521  & 0.761  \\
          & 192   & \textcolor[rgb]{ 1,  0,  0}{0.408 } & \textcolor[rgb]{ 1,  0,  0}{0.402 } & 0.434  & 0.430  & 0.437  & 0.416  & \textcolor[rgb]{ 0,  .439,  .753}{0.416 } & \textcolor[rgb]{ 0,  .439,  .753}{0.408 } & 0.543  & 0.684  & 0.457  & 0.471  & 0.568  & 0.598  \\
    \midrule
    \multirow{4}[2]{*}{\begin{sideways}ETTh2\end{sideways}} & 48    & \textcolor[rgb]{ 1,  0,  0}{0.291 } & \textcolor[rgb]{ 1,  0,  0}{0.192 } & \textcolor[rgb]{ 0,  .439,  .753}{0.301 } & 0.241  & \textcolor[rgb]{ 0,  .439,  .753}{0.301 } & \textcolor[rgb]{ 0,  .439,  .753}{0.223 } & 0.307  & 0.228  & 0.357  & 0.339  & 0.329  & 0.297  & 0.414  & 0.458  \\
          & 96    & \textcolor[rgb]{ 1,  0,  0}{0.333 } & \textcolor[rgb]{ 0,  .439,  .753}{0.302 } & 0.375  & 0.309  & \textcolor[rgb]{ 0,  .439,  .753}{0.350 } & 0.303  & \textcolor[rgb]{ 0,  .439,  .753}{0.350 } & \textcolor[rgb]{ 0,  .439,  .753}{0.295 } & 0.417  & 0.421  & 0.375  & 0.376  & 0.425  & 0.490  \\
          & 144   & \textcolor[rgb]{ 1,  0,  0}{0.361 } & \textcolor[rgb]{ 1,  0,  0}{0.335 } & \textcolor[rgb]{ 0,  .439,  .753}{0.368 } & \textcolor[rgb]{ 0,  .439,  .753}{0.338 } & 0.394  & 0.345  & 0.394  & 0.354  & 0.489  & 0.583  & 0.401  & 0.420  & 0.497  & 0.680  \\
          & 192   & \textcolor[rgb]{ 1,  0,  0}{0.382 } & \textcolor[rgb]{ 0,  .439,  .753}{0.377 } & \textcolor[rgb]{ 0,  .439,  .753}{0.388 } & \textcolor[rgb]{ 1,  0,  0}{0.353 } & 0.410  & 0.382  & 0.410  & 0.387  & 0.488  & 0.454  & 0.457  & 0.465  & 0.512  & 0.682  \\
    \midrule
    \multirow{4}[2]{*}{\begin{sideways}ETTm1\end{sideways}} & 48    & \textcolor[rgb]{ 1,  0,  0}{0.334 } & 0.315  & \textcolor[rgb]{ 0,  .439,  .753}{0.336 } & \textcolor[rgb]{ 1,  0,  0}{0.303 } & 0.338  & \textcolor[rgb]{ 1,  0,  0}{0.289 } & 0.356  & 0.323  & 0.395  & 0.412  & 0.354  & 0.348  & 0.529  & 0.670  \\
          & 96    & \textcolor[rgb]{ 1,  0,  0}{0.335 } & \textcolor[rgb]{ 1,  0,  0}{0.303 } & 0.347  & 0.307  & \textcolor[rgb]{ 0,  .439,  .753}{0.346 } & \textcolor[rgb]{ 0,  .439,  .753}{0.305 } & 0.347  & 0.312  & 0.406  & 0.492  & 0.353  & 0.318  & 0.482  & 0.624  \\
          & 144   & \textcolor[rgb]{ 1,  0,  0}{0.347 } & \textcolor[rgb]{ 1,  0,  0}{0.322 } & 0.359  & 0.341  & 0.362  & 0.320  & 0.360  & 0.326  & 0.466  & 0.562  & 0.367  & 0.345  & 0.512  & 0.683  \\
          & 192   & \textcolor[rgb]{ 1,  0,  0}{0.356 } & \textcolor[rgb]{ 1,  0,  0}{0.334 } & 0.294  & 0.347  & 0.375  & 0.344  & 0.363  & 0.337  & 0.454  & 0.533  & 0.376  & 0.361  & 0.531  & 0.591  \\
    \midrule
    \multirow{4}[2]{*}{\begin{sideways}ETTm2\end{sideways}} & 48    & \textcolor[rgb]{ 1,  0,  0}{0.230 } & \textcolor[rgb]{ 0,  .439,  .753}{0.147 } & 0.231  & 0.152  & 0.232  & \textcolor[rgb]{ 1,  0,  0}{0.135 } & 0.239  & 0.148  & 0.254  & 0.189  & 0.247  & 0.192  & 0.279  & 0.241  \\
          & 96    & \textcolor[rgb]{ 1,  0,  0}{0.252 } & 0.184  & 0.254  & 0.179  & \textcolor[rgb]{ 0,  .439,  .753}{0.253 } & \textcolor[rgb]{ 0,  .439,  .753}{0.172 } & 0.254  & \textcolor[rgb]{ 1,  0,  0}{0.170 } & 0.307  & 0.244  & 0.278  & 0.265  & 0.372  & 0.364  \\
          & 144   & \textcolor[rgb]{ 1,  0,  0}{0.272 } & 0.214  & 0.277  & 0.216  & 0.284  & \textcolor[rgb]{ 0,  .439,  .753}{0.213 } & \textcolor[rgb]{ 0,  .439,  .753}{0.276 } & \textcolor[rgb]{ 1,  0,  0}{0.201 } & 0.342  & 0.288  & 0.309  & 0.288  & 0.408  & 0.352  \\
          & 192   & \textcolor[rgb]{ 1,  0,  0}{0.287 } & \textcolor[rgb]{ 0,  .439,  .753}{0.223 } & 0.294  & 0.236  & 0.294  & 0.224  & \textcolor[rgb]{ 0,  .439,  .753}{0.291 } & \textcolor[rgb]{ 1,  0,  0}{0.220 } & 0.543  & 0.341  & 0.327  & 0.430  & 0.485  & 0.892  \\
    \midrule
    \multirow{4}[2]{*}{\begin{sideways}Exchange\end{sideways}} & 48    & \textcolor[rgb]{ 1,  0,  0}{0.139 } & \textcolor[rgb]{ 1,  0,  0}{0.045 } & 0.149  & \textcolor[rgb]{ 0,  .439,  .753}{0.046 } & \textcolor[rgb]{ 0,  .439,  .753}{0.145 } & 0.048  & \textcolor[rgb]{ 0,  .439,  .753}{0.145 } & \textcolor[rgb]{ 0,  .439,  .753}{0.046 } & 0.187  & 0.073  & 0.159  & 0.067  & 0.205  & 0.124  \\
          & 96    & \textcolor[rgb]{ 1,  0,  0}{0.195 } & \textcolor[rgb]{ 1,  0,  0}{0.083 } & 0.211  & 0.092  & \textcolor[rgb]{ 0,  .439,  .753}{0.204 } & 0.090  & 0.223  & \textcolor[rgb]{ 0,  .439,  .753}{0.089 } & 0.294  & 0.159  & 0.247  & 0.168  & 0.778  & 0.409  \\
          & 144   & \textcolor[rgb]{ 1,  0,  0}{0.241 } & \textcolor[rgb]{ 1,  0,  0}{0.113 } & 0.265  & 0.141  & 0.265  & 0.138  & \textcolor[rgb]{ 0,  .439,  .753}{0.256 } & \textcolor[rgb]{ 0,  .439,  .753}{0.133 } & 0.375  & 0.292  & 0.272  & 0.310  & 0.680  & 0.671  \\
          & 192   & \textcolor[rgb]{ 1,  0,  0}{0.284 } & \textcolor[rgb]{ 1,  0,  0}{0.180 } & 0.329  & 0.212  & \textcolor[rgb]{ 0,  .439,  .753}{0.298 } & \textcolor[rgb]{ 0,  .439,  .753}{0.181 } & 0.301  & 0.182  & 0.464  & 0.494  & 0.354  & 0.403  & 0.979  & 0.544  \\
    \midrule
    \multirow{4}[2]{*}{\begin{sideways}Electricity\end{sideways}} & 48    & \textcolor[rgb]{ .357,  .608,  .835}{0.247 } & \textcolor[rgb]{ 0,  .439,  .753}{0.155 } & 0.275  & 0.175  & 0.251  & \textcolor[rgb]{ 1,  0,  0}{0.143 } & \textcolor[rgb]{ 1,  0,  0}{0.242 } & 0.158  & 0.259  & 0.156  & 0.287  & 0.209  & 0.290  & 0.198  \\
          & 96    & \textcolor[rgb]{ 1,  0,  0}{0.243 } & \textcolor[rgb]{ 0,  .439,  .753}{0.152 } & 0.285  & 0.184  & \textcolor[rgb]{ 0,  .439,  .753}{0.244 } & \textcolor[rgb]{ 1,  0,  0}{0.144 } & 0.245  & 0.155  & 0.277  & 0.180  & 0.271  & 0.181  & 0.303  & 0.201  \\
          & 144   & \textcolor[rgb]{ 1,  0,  0}{0.243 } & \textcolor[rgb]{ 0,  .439,  .753}{0.151 } & 0.291  & 0.192  & \textcolor[rgb]{ 0,  .439,  .753}{0.244 } & \textcolor[rgb]{ 1,  0,  0}{0.147 } & 0.247  & 0.153  & 0.287  & 0.191  & 0.278  & 0.186  & 0.320  & 0.211  \\
          & 192   & \textcolor[rgb]{ 0,  .439,  .753}{0.245 } & \textcolor[rgb]{ 0,  .439,  .753}{0.152 } & 0.299  & 0.198  & \textcolor[rgb]{ 1,  0,  0}{0.241 } & \textcolor[rgb]{ 1,  0,  0}{0.144 } & 0.246  & 0.155  & 0.296  & 0.189  & 0.291  & 0.204  & 0.319  & 0.216  \\
    \midrule
    \multirow{4}[2]{*}{\begin{sideways}Weather\end{sideways}} & 48    & \textcolor[rgb]{ 0,  .439,  .753}{0.182 } & \textcolor[rgb]{ 0,  .439,  .753}{0.128 } & \textcolor[rgb]{ 1,  0,  0}{0.171 } & \textcolor[rgb]{ 1,  0,  0}{0.127 } & 0.183  & 0.137  & 0.200  & 0.155  & 0.218  & 0.159  & 0.185  & 0.136  & 0.242  & 0.187  \\
          & 96    & \textcolor[rgb]{ 1,  0,  0}{0.199 } & \textcolor[rgb]{ 1,  0,  0}{0.154 } & \textcolor[rgb]{ 0,  .439,  .753}{0.203 } & \textcolor[rgb]{ 0,  .439,  .753}{0.155 } & \textcolor[rgb]{ 0,  .439,  .753}{0.203 } & 0.158  & 0.223  & 0.180  & 0.222  & 0.161  & 0.206  & 0.159  & 0.283  & 0.230  \\
          & 144   & \textcolor[rgb]{ 1,  0,  0}{0.224 } & \textcolor[rgb]{ 1,  0,  0}{0.166 } & \textcolor[rgb]{ 0,  .439,  .753}{0.225 } & \textcolor[rgb]{ 0,  .439,  .753}{0.171 } & 0.226  & 0.175  & 0.241  & 0.192  & 0.238  & 0.178  & 0.227  & 0.172  & 0.310  & 0.307  \\
          & 192   & \textcolor[rgb]{ 1,  0,  0}{0.240 } & \textcolor[rgb]{ 1,  0,  0}{0.192 } & \textcolor[rgb]{ 0,  .439,  .753}{0.241 } & \textcolor[rgb]{ 0,  .439,  .753}{0.193 } & \textcolor[rgb]{ 0,  .439,  .753}{0.241 } & 0.197  & 0.257  & 0.211  & 0.254  & 0.199  & 0.243  & 0.196  & 0.355  & 0.344  \\
    \bottomrule
    \end{tabular}%
  \label{tab:baseline}%
  }
\end{table*}%

\section{Experiments}
\label{sec:experiments}
  
\subsection{Datasets and Baselines}
  
We conduct comprehensive experiments to assess the performance of TimeSieve. For multivariate forecasting, we utilize a variety of real-world benchmarks, including ETT \cite{zhou2021informer}, Exchange \cite{lai2018modeling}, Electricity \cite{lin2023segrnn}, and Weather \cite{zeng2022transformers}. Notably, instead of using a fixed lookback window length, we follow the approach of Koopa \cite{liu2023koopa} by setting the lookback window length $T$ to twice the forecast window length (i.e., $T=2H$). This approach is based on the fact that in real-world scenarios, historical observations are readily accessible, and utilizing a larger amount of observed data can enhance the performance of deep models, particularly as the forecast horizon extends \cite{liu2023koopa}. 

We extensively compare TimeSieve with state-of-the-art deep forecasting models, including Transformer-based models such as PatchTST \cite{nie2023time}, Nonstationary-Transformer (NSTformer) \cite{liu2023nonstationary} and Autoformer \cite{wu2021autoformer}; MLP-based models such as LightTS \cite{campos2023lightts}, DLinear \cite{zeng2022transformers} and Koopman operator-based models such as Koopa \cite{liu2023koopa}. See the \textbf{Appendix Implementation Details} for a more detailed introduction.


\subsection{Comparative Experiment}
  
We present the forecasting results in Table \ref{tab:baseline}, where \textbf{\textcolor{red}{Red}}
indicates the best result and  \textcolor[rgb]{ 0,  .439,  .753}{Blue} refers to the second best. Across Table \ref{tab:baseline}, TimeSieve consistently outperformed the baseline models in most scenarios, particularly excelling in datasets with  redundant features, such as ETTh1 and ETTh2. In the ETTh1 dataset, TimeSieve achieved a significant reduction in both MAE and MSE at all prediction horizons. At the 48-time step horizon in ETTh1, TimeSieve achieved an MAE of 0.361 and an MSE of 0.341, outperforming Koopa by approximately 6.2\% in MAE and 6.3\% in MSE. This consistent performance can be attributed to the effective integration of the Wavelet Decomposition Block (WDB) and the Information Filtering and Compression Block (IFCB), which together capture multi-scale features while filtering out redundant features. However, it did not achieve the best results in some experiments with a forecast window of 48. We believe this may be due to the significant short-term fluctuations in time series data, making it difficult for the model to distinguish meaningful signals from redundant features in shorter forecast windows. In contrast, longer windows (e.g., 144, 192) provide more historical data, enabling the model to better capture long-term dependencies and seasonal patterns, and to differentiate redundant features from short-term features.

\begin{table}[tbp!]
  \centering
  \caption{Comparison of sequence decomposition methods with and without wavelet transformations (WDB \& WRB and No Wavelet) on Exchange and ETTh2 datasets. We set the lookback length T = 2H.}
  \begin{tabular}{c|ccccc}
    \toprule
    \multicolumn{2}{c}{\textbf{Method}} & \multicolumn{2}{c}{\textbf{No Wavelet}} & \multicolumn{2}{c}{\textbf{WDB\&WRB}} \\
    \multicolumn{2}{c}{\textbf{Metric}} & \textbf{MAE}\small{$\downarrow$}    & \textbf{MSE}\small{$\downarrow$}    & \textbf{MAE}\small{$\downarrow$}    & \textbf{MSE}\small{$\downarrow$}  \\
    \midrule
    \multirow{4}[2]{*}{\begin{sideways}\textbf{Exchange}\end{sideways}} 
          & 48    & 0.151  & 0.050  & \textbf{0.139}  & \textbf{0.045}  \\
          & 96    & 0.204  & 0.093  & \textbf{0.195}  & \textbf{0.083}  \\
          & 144   & 0.261  & 0.202  & \textbf{0.241}  & \textbf{0.113}  \\
          & 192   & 0.312  & 0.251  & \textbf{0.284}  & \textbf{0.180}  \\
    \midrule
    \multirow{4}[2]{*}{\begin{sideways}\textbf{ETTh2}\end{sideways}} 
          & 48    & 0.299  & 0.246  & \textbf{0.293}  & \textbf{0.193}  \\
          & 96    & 0.342  & 0.333  & \textbf{0.337}  & \textbf{0.308}  \\
          & 144   & 0.370  & 0.374  & \textbf{0.363}  & \textbf{0.365}  \\
          & 192   & 0.389  & 0.402  & \textbf{0.385}  & \textbf{0.387}  \\
          \midrule
    \multirow{4}[2]{*}{\begin{sideways}\textbf{ETTm1}\end{sideways}} 
          & 48    & 0.339  & 0.331  & \textbf{0.334}  & \textbf{0.315}  \\
          & 96    & 0.342  & 0.341  & \textbf{0.335}  & \textbf{0.303}  \\
          & 144   & 0.359  & 0.347  & \textbf{0.347}  & \textbf{0.322}  \\
          & 192   & 0.360  & 0.352  & \textbf{0.356}  & \textbf{0.334}  \\
    \bottomrule
    \end{tabular}%
  \label{tab:WRB}%
\end{table}%


Interestingly, the Exchange dataset showed the most significant relative improvement. At the 48-time step horizon, TimeSieve achieved an MAE of 0.139 and an MSE of 0.045, outperforming Koopa by 6.7\% in MAE and 2.2\% in MSE. 

In analysis, we observed an interesting phenomenon where the model's MSE performance is worse compared to MAE. This can be attributed to MSE's sensitivity to outliers and large errors, as it squares the errors before averaging them. This sensitivity may impact performance, particularly because the model may filter out significant extreme values. This effect is especially pronounced in the ETTm2 dataset, which is characterized by a higher sampling frequency and an operational environment that may be more susceptible to extreme values. The formulas for calculating MSE and MAE are provided in the \textbf{Appendix Metric}.

The varying performance across different datasets and prediction horizons (48, 96, 144, 192 steps) underscores the model's versatility. TimeSieve consistently demonstrated substantial performance gains at longer horizons (144 and 192 steps). This is particularly evident in the ETTm1 and ETTh1 datasets, where longer-term predictions are crucial for timely decision-making. For example, in the ETTh1 dataset at the 192-step horizon, TimeSieve achieved an MAE of 0.408 and an MSE of 0.402, showcasing substantial improvements over other models such as Koopa (MAE 0.434, MSE 0.430) and PatchTST (MAE 0.437, MSE 0.416). Similarly, in the ETTm1 dataset at the 192-step horizon, the model achieved an MAE of 0.356 and an MSE of 0.334, significantly outperforming competitors in terms of both metrics. The integrated wavelet transform and information bottleneck approach effectively balance feature extraction and redundant feature compression, ensuring a certain level of reliable predictions even when data is limited.

In conclusion, TimeSieve's ability to achieve state-of-the-art performance on the majority of the datasets tested highlights its superior predictive accuracy and generalization capability.

\subsection{Ablation Study}
  
To validate the effectiveness of the modules used in our proposed model, we conducted ablation studies focusing on the wavelet transformation and the IFCB components. Further discussions on the selection of wavelet basis functions and the impact of \(\mathcal{L}_{IB}\) loss weights can be found in \textbf{Appendix Tables}.

\noindent \textbf{Wavelet Transformation. }

\begin{table*}[tbp]
    
  \centering
  \caption{Comparison of IFCB performance on Exchange and ETTh2 datasets under different conditions. $Only A$ applies IFCB only on the approximation coefficients after WDB, $Only D$ applies IFCB only on the detail coefficients after WRB, and 'TimeSieve' applies IFCB on all coefficients. }
    \begin{tabular}{c|ccccccc}
    \toprule
    \multicolumn{2}{c}{\textbf{Models}} & \multicolumn{2}{c}{\textbf{TimeSieve}} & \multicolumn{2}{c}{\textbf{$Only A$}} & \multicolumn{2}{c}{\textbf{$Only D$}} \\
    \multicolumn{2}{c}{\textbf{Metric}} & \textbf{MAE}\small{$\downarrow$}   & \textbf{MSE}\small{$\downarrow$}   & \textbf{MAE}\small{$\downarrow$}   & \textbf{MSE}\small{$\downarrow$}   & \textbf{MAE}\small{$\downarrow$}   & \textbf{MSE}\small{$\downarrow$} \\
    \midrule
    \multirow{4}[2]{*}{\begin{sideways}    \textbf{Exchange}\end{sideways}} 
          & 48    & \textbf{0.139}  & \textbf{0.045}  & 0.141  & 0.048  & \textbf{0.139}  & 0.047  \\
          & 96    &\textbf{ 0.195}  &\textbf{0.083} & 0.198  & 0.084  & \textbf{ 0.195}  & \textbf{0.083}  \\
          & 144   & \textbf{0.241}  & \textbf{0.113}  & 0.245  & 0.125  & 0.245  & 0.132  \\
          & 192   & \textbf{0.284}  & \textbf{0.180}  & 0.288  & 0.215  & 0.290  & 0.188  \\
    \midrule
    \multirow{4}[2]{*}{\begin{sideways}  \textbf{ETTh2}\end{sideways}} 
          & 48    & \textbf{0.291}  & \textbf{0.192}  & 0.293  & 0.200  & 0.293  & 0.239  \\
          & 96    & \textbf{0.333}  & \textbf{0.302}  & 0.337  & 0.311  & 0.338  & 0.312  \\
          & 144   & \textbf{0.361}  & \textbf{0.335}  & 0.363  & 0.389  & 0.369  & 0.389  \\
          & 192   & \textbf{0.382}  & \textbf{0.377}  & 0.385   & 0.408  & 
            0.385   & 0.406  \\
    \bottomrule
    \end{tabular}%
  \label{tab:IB}%
      
\end{table*}%
\begin{figure}[tbp]
    
  \centering
  \includegraphics[width=\linewidth]{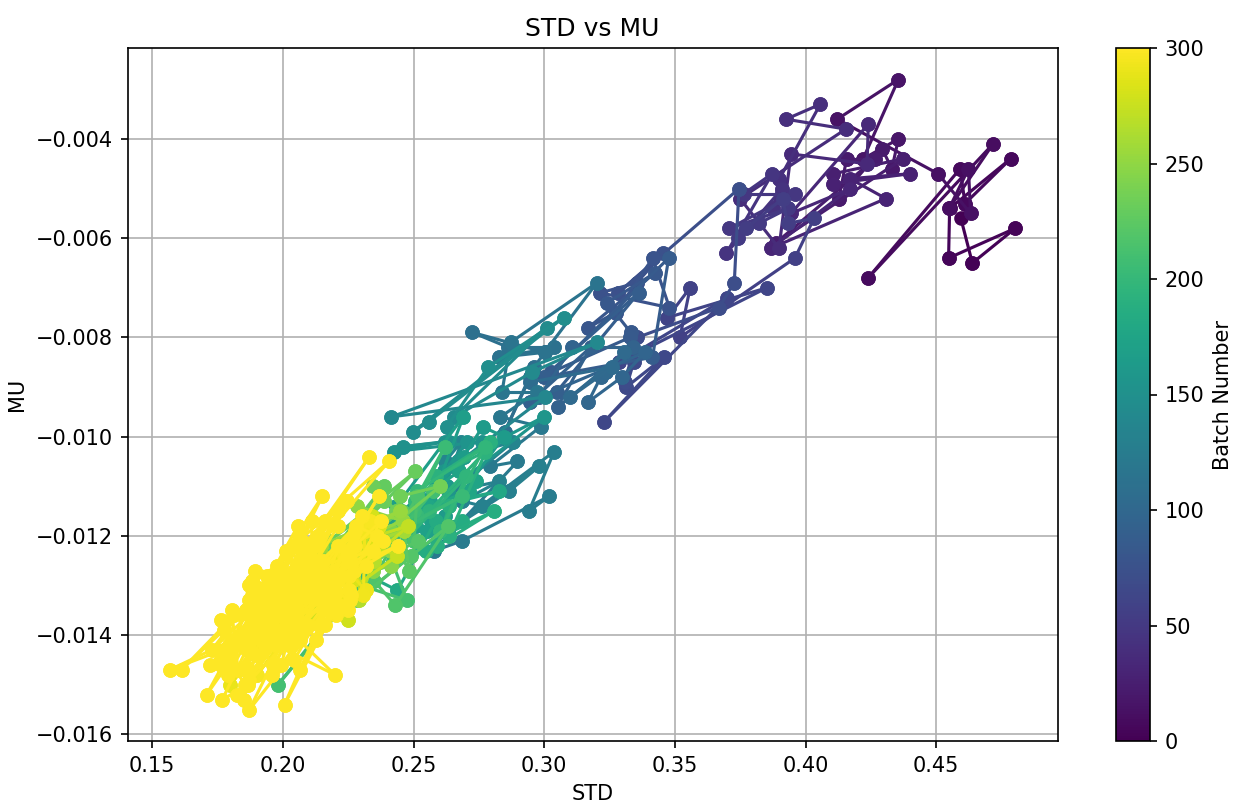}
  \caption{The figure depicts the evolution of the mean (MU) and standard deviation (STD) within the information bottleneck as model iterations progress, with MSE as the metric, for a prediction length of 48 on the Exchange dataset.}
  \label{fig:MUSTD}
      
\end{figure}
\begin{figure}[tbp]
    
  \centering
  \includegraphics[width=\linewidth]{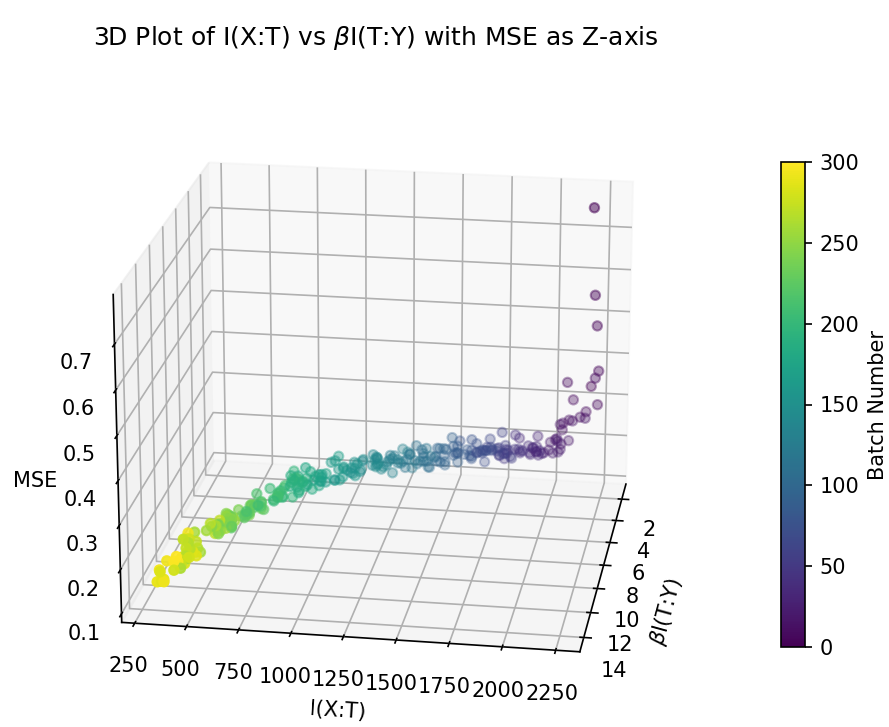}
  \caption{The figure illustrates the changes in the mutual information related to the input \( I(X;T) \) and the output \(\beta I(T;Y)\) throughout model iterations, measured by MSE, for a prediction length of 48 on the Exchange dataset.}
  \label{fig:IXTY}
      
\end{figure}

In Table \ref{tab:WRB}, we compare the performance of our sequence decomposition methods with and without wavelet transformations(WDB\&WRB and No Wavelet)  on the Exchange,  ETTh2 and ETTm1 datasets. The results demonstrate that incorporating wavelet transformations significantly improves forecasting accuracy, as evidenced by lower MAE and MSE values across all forecasting horizons.

By leveraging wavelet transformations, our model can decompose the time series data into different frequency components, effectively compressing redundant features. This process enhances the model's ability to capture the true underlying trends and patterns, leading to significant improvements in both MAE and MSE. For example, at the 96-step horizon on the Exchange dataset, the MAE decreases from 0.204 to 0.195 and the MSE decreases from 0.093 to 0.083. Similarly, on the ETTh2 dataset, the MAE decreases from 0.342 to 0.337 and the MSE decreases from 0.333 to 0.308.

The results presented in Table \ref{tab:WRB} validate the effectiveness of using wavelet transformations for time series decomposition. By leveraging the multi-scale nature of wavelets, our method can effectively capture both the coarse and fine-grained structures in the data, leading to more accurate and reliable forecasts.

\noindent \textbf{Information Bottleneck }

In Table \ref{tab:IB}, we present a comparison of our proposed TimeSieve method against two baseline methods, $Only A$ and $Only D$, on the Exchange and ETTh2 datasets. Results show that TimeSieve consistently outperforms the baseline methods across all forecasting horizons. For instance, at a 96-step forecasting horizon on the ETTh1 dataset, TimeSieve achieved a MAE of 0.333 and an MSE of 0.302, which is superior to $Only A$ (MAE 0.337, MSE 0.311) and $Only D$ (MAE 0.338, MSE 0.312). However, improvements are more pronounced in the Exchange dataset. In shorter forecast horizons at Exchange, the enhancements are not as apparent, which may be due to the dataset's richer set of influential features, making it challenging for IFCB to discern effectively at shorter time steps.

Figure \ref{fig:MUSTD} presents a scatter plot illustrating the evolution of the standard deviation and mean values in the information bottleneck as the model iterates. In this study, by observing the changes in the mean and standard deviation throughout the training process, we have identified a significant decreasing trend over time, with both parameters eventually converging within a more defined range. The convergence of mu suggests that the model's latent representations are progressively approaching the central distribution of the data, which aids the model in identifying more representative and universally applicable features across the entire sample space. The reduction in std likely indicates an enhancement in the predictability and certainty of the model's outputs.

Figure \ref{fig:IXTY} presents a scatter plot of the mutual information values in the objective function as the model iterates. In this study, we observed that during the model training process, the mutual information related to the input \( I(X;T) \) gradually decreases with increasing iterations, while the mutual information related to the output \( I(T;Y) \) correspondingly increases, alongside a consistent reduction in MSE. This trend validates the effectiveness of the Information Bottleneck (IB) theory in practical applications, demonstrating that the model compresses input information while increasingly retaining information crucial for predicting the output. Specifically, the reduction in \( I(X;T) \) reflects the model’s effective filtration of input data during the learning process. Concurrently, the increase in \( I(T;Y) \) suggests an enhancement in the model’s capability to extract features closely associated with the task, thereby improving predictive accuracy.

\section{Conclusions}
\label{sec:conclusions}
In this paper, we presented the TimeSieve framework for time series forecasting, emphasizing the decomposition and reconstruction of time series data through wavelet and information bottleneck techniques. Our approach integrates Wavelet Decomposition Block, Wavelet Reconstruction Block, and Information Fusion and Compression Block to enhance prediction accuracy.

We conducted extensive experiments on multiple benchmark datasets, demonstrating the superiority of TimeSieve in comparison to state-of-the-art methods. The results highlight the model's capacity to handle varying forecast lengths and different types of time series data.

The proposed TimeSieve framework establishes a foundational approach for time series forecasting across various domains, including finance and climate science. Future work will focus on enhancing the model’s robustness against datasets, alongside exploring extending the model to handle  multi-modal time series data.



\bibliographystyle{unsrt}  
\bibliography{references}

\appendix
\label{MCT}



\section{Implementation Details}
\label{ID}
\subsection{Baselines}
\label{baseline}
\subsubsection{Koopa \cite{liu2023koopa}}
Koopa is a novel forecasting model designed to handle the intrinsic non-stationarity in real-world time series. Developed by researchers at Tsinghua University, Koopa leverages modern Koopman theory to address the complexities associated with changing temporal distributions. The model decomposes intricate non-stationary series into time-variant and time-invariant components using a Fourier Filter. Subsequently, it employs specialized Koopman Predictors to advance the dynamics of these components independently.

The innovative architecture of Koopa involves stackable blocks that learn hierarchical dynamics, facilitating a modular approach to time series forecasting. Each block integrates Koopman embedding and utilizes Koopman operators as linear representations of implicit transitions. For time-variant dynamics, the model calculates context-aware operators within the temporal neighborhood, ensuring adaptability to localized changes. This method enables Koopa to effectively extend the forecast horizon by incorporating incoming ground truth data.

\subsubsection{PatchTST \cite{nie2023time}}
PatchTST is a Transformer-based model designed for efficient and accurate long-term time series forecasting. Developed by researchers at Princeton University and IBM Research, PatchTST introduces two key innovations: the segmentation of time series into subseries-level patches and a channel-independent approach where each channel corresponds to a single univariate time series. These patches serve as input tokens to the Transformer, enabling the model to retain local semantic information, reduce the computational complexity of attention mechanisms, and effectively utilize longer historical sequences.

The model's architecture leverages the benefits of patching to enhance locality in the embeddings and minimize the quadratic increase in computation and memory usage. The channel-independent design allows PatchTST to achieve significant improvements in forecasting accuracy by sharing the same embedding and Transformer weights across all channels while processing them independently. PatchTST has demonstrated superior performance in both supervised and self-supervised learning tasks, outperforming state-of-the-art Transformer-based models in various benchmark datasets.

\subsubsection{DLinear \cite{zeng2022transformers}}
DLinear is a simple yet effective model designed to challenge the dominance of Transformer-based models in long-term time series forecasting (LTSF). Developed by researchers at the Chinese University of Hong Kong and the International Digital Economy Academy, DLinear is built on the premise that complex architectures like Transformers may not be necessary for capturing temporal dependencies in time series data.

DLinear operates by decomposing the time series into a trend component and a remainder component. It then applies two separate one-layer linear networks to model these components independently. This direct multi-step (DMS) forecasting approach contrasts with the autoregressive methods commonly used in traditional models, which can suffer from error accumulation over long prediction horizons.

Despite its simplicity, DLinear has demonstrated superior performance across various benchmarks, often surpassing Transformer-based models. Its strengths lie in its ability to capture both short-term and long-term dependencies efficiently, with a low computational cost and high interpretability. DLinear's success raises important questions about the necessity of using complex Transformer architectures for LTSF and suggests a reevaluation of their application in time series forecasting tasks.

\subsubsection{NSTformer \cite{liu2023nonstationary}}
The Nonstationary Transformer (NTSformer) is an advanced framework designed to address the challenges posed by non-stationary time series data in forecasting tasks. Developed by researchers at Tsinghua University, this model builds upon the conventional Transformer architecture, which excels at capturing temporal dependencies through its attention mechanism but struggles with non-stationary data where statistical properties and joint distributions change over time.

To mitigate the limitations of standard Transformers, the Nonstationary Transformer introduces two key modules: Series Stationarization and De-stationary Attention. Series Stationarization employs a normalization strategy to align the statistical properties of input time series, enhancing predictability by ensuring a more stable data distribution. This module normalizes the input series to a common scale, reducing the discrepancies caused by varying means and variances.

However, purely stationarizing the series can lead to the over-stationarization problem, where the model fails to capture crucial temporal dependencies intrinsic to non-stationary data. To address this, the De-stationary Attention module reintegrates the non-stationary characteristics into the attention mechanism. This module learns de-stationary factors that adjust the attention weights, preserving the unique temporal patterns of the original series.

\subsubsection{LightTS \cite{campos2023lightts}}
LightTS is an innovative framework designed to perform lightweight time series classification, addressing the computational constraints often encountered in resource-limited environments such as edge devices. Developed by researchers from Aalborg University and several collaborating institutions, LightTS utilizes adaptive ensemble distillation to compress large, resource-intensive ensemble models into compact, efficient classifiers without compromising on accuracy.

The core of LightTS involves two primary innovations: adaptive ensemble distillation and Pareto optimal model selection. Adaptive ensemble distillation assigns dynamic weights to different base models, tailoring their contributions based on their classification capabilities, thereby distilling the ensemble knowledge into a single lightweight model. This approach ensures that the most relevant information is captured efficiently, leading to a high-performing student model. Furthermore, LightTS employs a multi-objective Bayesian optimization method to identify Pareto optimal settings that balance model accuracy and size, enabling users to select the most accurate lightweight models that fit within specific storage constraints.

\subsubsection{Autoformer \cite{wu2021autoformer}}
Autoformer is a framework designed to tackle the complexities of long-term time series forecasting. Developed by researchers at Tsinghua University, Autoformer innovatively combines decomposition architecture with an Auto-Correlation mechanism to enhance the model's ability to handle intricate temporal patterns. Traditional Transformer-based models often struggle with long-term dependencies due to the prohibitive computational costs and the obscured nature of temporal dependencies in extended forecasting horizons. Autoformer addresses these challenges by embedding series decomposition blocks within the deep learning architecture, allowing for progressive extraction and refinement of long-term trends and seasonal components.

The Auto-Correlation mechanism is a key innovation of Autoformer, inspired by stochastic process theory. This mechanism leverages the inherent periodicity in time series data to discover dependencies and aggregate similar sub-series efficiently. By replacing the self-attention mechanism with Auto-Correlation, Autoformer achieves a significant reduction in computational complexity, scaling as O(L logL) for a series of length L. This enables more efficient and accurate long-term forecasting compared to previous models.

\subsection{Datasets}
\label{dataset}
\textbf{Exchange Rate Dataset:} The Exchange Rate dataset consists of the daily exchange rates of eight different countries, spanning from 1990 to 2016. This dataset is commonly used to study and forecast currency fluctuations, providing valuable insights for economic planning and financial analysis. The data includes exchange rates for currencies such as USD, GBP, EUR, JPY, and others.

\noindent \textbf{ETT (Electricity Transformer Temperature) Dataset:} The ETT dataset includes data collected from electricity transformers, capturing both load and oil temperature measurements at 15-minute intervals over a period from July 2016 to July 2018. This dataset is particularly useful for energy consumption forecasting and transformer health monitoring. The dataset is split into ETTh1, ETTh2, ETTm1, and ETTm2 subsets, representing different transformer units and sampling frequencies (hourly and 15-minutely).

\noindent \textbf{Electricity Consumption Dataset:} The Electricity Consumption dataset contains hourly electricity usage data of 321 clients from 2012 to 2014. This dataset is utilized to predict future electricity demand and helps in managing energy resources efficiently. The data captures variations in electricity usage patterns, which are influenced by factors such as weather, economic conditions, and consumer behavior.

\noindent \textbf{Weather dataset} The Weather dataset, provided by the Max Planck Institute for Biogeochemistry, comprises detailed meteorological measurements from 2020 to 2021. Recorded every ten minutes, this dataset includes variables such as temperature, humidity, and wind speed. It is essential for high-resolution weather forecasting and climate research, offering insights into microclimatic fluctuations and seasonal patterns. This rich dataset serves as a valuable resource for understanding and predicting weather-related changes over short intervals.

\begin{table}[htbp]
  \centering
  \caption{Summary of Datasets Used in Time Series Forecasting}
  \begin{tabular}{|c|c|c|c|}
    \hline
    \textbf{Dataset} & \textbf{Time Points} & \textbf{Frequency} & \textbf{Variables} \\
    \hline
    ETTh1 & 17420 & Hour & 7 \\
    \hline
    ETTh2 & 17420 & Hour & 7 \\
    \hline
    ETTm1 & 69680 & 15 Min & 7 \\
    \hline
    ETTm2 & 69680 & 15 Min & 7 \\
    \hline
    Exchange Rate & 7588 & Day & 8 \\
    \hline
    Electricity & 26304 & Hour & 321 \\
    \hline
    Weather  & 52695 & 10Min & 21 \\
    \hline
  \end{tabular}
  \label{tab:dataset_summary}
\end{table}
\subsection{Definition of Metric}
\label{metric}
\textbf{Mean Absolute Error (MAE)}: The Mean Absolute Error is a metric used to measure the average magnitude of errors between predicted values and actual values. It is defined as follows:

\begin{equation}
\text{MAE} = \frac{1}{n} \sum_{i=1}^{n} \left| y_i - \hat{y}_i \right|
\end{equation}

\noindent where \( y_i \) represents the actual value, \( \hat{y}_i \) represents the predicted value, and \( n \) is the number of observations.

\noindent \textbf{Mean Squared Error (MSE)}: The Mean Squared Error is another metric used to measure the average of the squares of the errors between predicted values and actual values. It is defined as follows:

\begin{equation}
\text{MSE} = \frac{1}{n} \sum_{i=1}^{n} \left( y_i - \hat{y}_i \right)^2
\end{equation}

where \( y_i \) represents the actual value, \( \hat{y}_i \) represents the predicted value, and \( n \) is the number of observations.

\subsection{Experimental Setup}
\begin{table}[h]
    \centering
    \begin{tabular}{|l|l|}
        \hline
        \textbf{Component}      & \textbf{Description}                       \\ \hline
        \textbf{GPU}            & NVIDIA GeForce RTX 3080                    \\ \hline
        \textbf{CPU}            & Intel(R) Xeon(R) E5-2686 v4                \\ \hline
        \textbf{Epochs}         & 10                                         \\ \hline
        \textbf{Evaluation Metrics}     & MAE and MSE                         \\ \hline
        \textbf{Batch Size}     & 32                                         \\ \hline
        \textbf{Optimizer}      & Adam                                 
             \\ \hline
        \textbf{Learning Rate}  & 0.0001                                     \\ \hline
        \textbf{Framework}      & PyTorch 2.0.1                              \\ \hline
        \textbf{Amount of Memory}      & 32G                              \\ \hline
        \textbf{CUDA Version}   & CUDA 11.8                                  \\ \hline
    \end{tabular}
    \caption{Experimental Setup: Summary of Hardware and Software Configurations}
    \label{tab:hardware_software}
\end{table}



\section{Comparison of of different \(\mathcal{L}_{IB}\) loss weights}
\begin{table*}[htbp]
  \hspace{-1cm}
  \centering
  \caption{Comparison of different \(\mathcal{L}_{IB}\) loss weights on Exchange rate dataset. The lookback length is set to $T = 2H$.}
  \begin{tabular}{c|cc|cc|cc|cc|cc|cc}
    \toprule
    \textbf{IB Weight} & \multicolumn{2}{c}{0.000001} & \multicolumn{2}{c}{0.00001} & \multicolumn{2}{c}{0.001} & \multicolumn{2}{c}{0.01} & \multicolumn{2}{c}{0.1}  \\
    \textbf{Metric} & \textbf{MAE}\small{$\downarrow$} & \textbf{MSE}\small{$\downarrow$} & \textbf{MAE}\small{$\downarrow$} & \textbf{MSE}\small{$\downarrow$} & \textbf{MAE}\small{$\downarrow$} & \textbf{MSE}\small{$\downarrow$} & \textbf{MAE}\small{$\downarrow$} & \textbf{MSE}\small{$\downarrow$} & \textbf{MAE}\small{$\downarrow$} & \textbf{MSE}\small{$\downarrow$} &  \\
    \midrule
    48 & 0.1394 & 0.0505 & 0.1391 & 0.0451 & 0.1390 & 0.0448 & \textbf{0.1389} & 0.0441 & 0.1390 & \textbf{0.0440}  \\
    96 & 0.1947 & 0.1102 & 0.1950 & \textbf{0.0833} & \textbf{0.1946} & 0.0834 & 0.1958 & 0.0857 & 0.1950 & 0.0854  \\
    144 & 0.2491 & 0.1392 & 0.2414 & 0.1333 & 0.2413 & \textbf{0.1133} & 0.2411 & 0.1535 & \textbf{0.2403} & 0.1527  \\
    192 & 0.3008 & 0.1886 & 0.2840 & 0.1802 & 0.2836 & 0.1803 & \textbf{0.2834} & \textbf{0.1778} & 0.2893 & 0.1804 \\
    \bottomrule
  \end{tabular}
  \label{tab:IB_weight}
\end{table*}
Table \ref{tab:IB_weight} presents the impact of varying information bottleneck (\(\mathcal{L}_{IB}\)) loss weights on model performance across different forecast horizons on the Exchange rate dataset. Notably, at a 48-step forecast, the lowest MAE and MSE were observed at IB weights of 0.01 and 0.1, respectively, suggesting that moderate IB weights help minimize predictive errors while maintaining accuracy.

At a 96-step forecast, the optimal MAE appeared at an IB weight of 0.001, while the lowest MSE was at 0.00001, indicating that smaller IB weights can sustain lower errors over longer forecasts.

For longer forecasts (144 and 192 steps), similar trends were observed with the lowest MAE and MSE appearing at adjusted IB weights, indicating that the selection of appropriate IB weights significantly influences model performance. However, when the IB weight is reduced beyond a certain threshold, the performance improvements plateau, suggesting that overly small IB weights might lose their effectiveness. 

\section{Comparison of wavelet basis functions}
\begin{table*}[htbp]
  \centering
  \caption{Comparison of different wavelet basis functions on Exchange rate dataset. The lookback length is set to $T = 2H$.}
  \begin{tabular}{c|cc|cc|cc}
    \toprule
    \textbf{Metric} & \multicolumn{2}{c}{\textbf{HAAR}} & \multicolumn{2}{c}{\textbf{SYM2}} & \multicolumn{2}{c}{\textbf{DB1}} \\
    & \textbf{MAE}\small{$\downarrow$} & \textbf{MSE}\small{$\downarrow$} & \textbf{MAE}\small{$\downarrow$} & \textbf{MSE}\small{$\downarrow$} & \textbf{MAE}\small{$\downarrow$} & \textbf{MSE}\small{$\downarrow$} \\
    \midrule
    48 & 0.1392 & 0.0478 & 0.1420 & 0.0450 & \textbf{0.1390} & \textbf{0.0448}  \\
    96 & \textbf{0.1945} & 0.1024 & 0.1945 & 0.0895 & 0.1946 & \textbf{0.0834} \\
    144 & 0.2446 & 0.1230 & 0.2430 & 0.1326 & \textbf{0.2413} & \textbf{0.1133}  \\
    192 & 0.3007 & 0.2840 & 0.2914 & 0.1804 & \textbf{0.2840} & \textbf{0.1803} \\
    \bottomrule
  \end{tabular}
  \label{tab:wavelet}
\end{table*}
Table \ref{tab:wavelet}showcases the comparative performance of different wavelet basis functions—HAAR, SYM2, and DB1—on the Exchange rate dataset with a lookback length set to \(T = 2H\). The performance metrics used are MAE and MSE, which provide insights into the accuracy and reliability of the predictions at various forecast horizons.

For a 48-step forecast, DB1 demonstrates the lowest MAE and MSE, indicating its superior ability to capture the underlying patterns in the data at shorter horizons. HAAR, while slightly less effective than DB1 in terms of MSE, offers comparable MAE values, suggesting that it remains a robust choice for certain types of financial data.

At a 96-step forecast, HAAR shows the lowest MAE, matching closely with SYM2, but DB1 outperforms both in terms of MSE, highlighting its efficiency in reducing prediction errors over longer horizons.

For 144 and 192-step forecasts, DB1 consistently shows the best performance in both MAE and MSE, underlining its effectiveness in handling longer-term dependencies in the dataset. SYM2, while competitive, tends to exhibit slightly higher errors, particularly in MSE, suggesting that its utility may be more limited under conditions requiring high precision.

Overall, the selection of the appropriate wavelet basis function significantly impacts model performance, with DB1 generally offering the best balance of accuracy and error reduction across the board. This analysis underscores the importance of choosing the right wavelet transform to enhance predictive performance in time series forecasting, especially in the context of financial datasets.


\end{document}